\documentclass[ijoc,sglanonrev]{informs4}
\usepackage{eqndefns-left} 
\RequirePackage{tgtermes}
\RequirePackage{newtxtext}
\RequirePackage{newtxmath}
\RequirePackage{bm}
\RequirePackage{endnotes}

\OneAndAHalfSpacedXII 

\usepackage{algorithm}
\usepackage{algpseudocode}
\usepackage{tikz}
\usepackage{multirow}
\usepackage{booktabs}
\usepackage{amsmath}
\usepackage{rotating}

\usepackage{natbib}
 \bibpunct[, ]{(}{)}{,}{a}{}{,}%
 \def\bibfont{\small}%
\usepackage{xcolor}

\usepackage{listings}
\usepackage{xcolor}
\usepackage{subcaption}
\usepackage{graphicx}   
\usepackage{hyperref}
\usepackage{array}

\usepackage{amssymb}
\usepackage{pifont}
\newcommand{\cmark}{{\textnormal{\ding{51}}}}%
\newcommand{\xmark}{{\textnormal{\ding{55}}}}%

\lstdefinestyle{pythonstyle}{
    language=Python,
    basicstyle=\ttfamily\footnotesize,       
    backgroundcolor=\color{gray!10},  
    keywordstyle=\color{blue}\bfseries,
    stringstyle=\color{green!50!black},
    commentstyle=\color{gray},
    numbers=left,                     
    numberstyle=\tiny\color{gray},
    stepnumber=1,
    frame=single,                     
    showstringspaces=false            
}

\EquationsNumberedThrough    

\TheoremsNumberedThrough     
\ECRepeatTheorems  %

\MANUSCRIPTNO{} 

\begin{document}


\RUNAUTHOR{Reijnen et al.}

\RUNTITLE{Job Shop Scheduling Benchmark}

\TITLE{Job Shop Scheduling Benchmark: Environments
and Instances for Learning and Non-learning Methods}

\ARTICLEAUTHORS{%
\AUTHOR{Robbert Reijnen\textsuperscript{a}, Igor G. Smit\textsuperscript{b}, Hongxiang Zhang\textsuperscript{c}, Yaoxin Wu\textsuperscript{a}, Zaharah Bukhsh\textsuperscript{a}, Yingqian Zhang\textsuperscript{a}}
\AFF{\textsuperscript{a} Department of Industrial Engineering \& Innovation Sciences, Eindhoven University of Technology,  \EMAIL{r.v.j.reijnen@tue.nl, y.wu2@tue.nl, z.bukhsh@tue.nl, yqzhang@tue.nl}}
\AFF{\textsuperscript{b} Department of Mathematics \& Computer Science, Eindhoven University of Technology,  \EMAIL{i.g.smit@tue.nl}}
\AFF{\textsuperscript{c} School of Transportation and Logistics, Southwest Jiaotong University, \EMAIL{hongxiang@my.swjtu.edu.cn}}
}

\ABSTRACT{
Job shop scheduling problems address the routing and sequencing of tasks in a job shop setting. Despite significant interest from operations research and machine learning communities over the years, a comprehensive platform for testing and comparing solution methods has been notably lacking. To fill this gap, we introduce a unified implementation of job shop scheduling problems and their solution methods, addressing the long-standing need for a standardized benchmarking platform in this domain. Our platform supports classic Job Shop (JSP), Flow Shop (FSP), Flexible Job Shop (FJSP), and Assembly Job Shop (AJSP), as well as variants featuring Sequence-Dependent Setup Times (SDST), variants with online arrivals of jobs, and combinations of these problems (e.g., FJSP-SDST and FAJSP). The platfrom provides a wide range of scheduling solution methods, from heuristics, metaheuristics, and exact optimization to deep reinforcement learning. The implementation is available as an open-source GitHub repository\footnote{\url{https://github.com/ai-for-decision-making-tue/Job_Shop_Scheduling_Benchmark_Environments_and_Instances}}, serving as a collaborative hub for researchers, practitioners, and those new to the field. Beyond enabling direct comparisons with existing methods on widely studied benchmark problems, this resource serves as a robust starting point for addressing constrained and complex problem variants. By establishing a comprehensive and unified foundation, this platform is designed to consolidate existing knowledge and to inspire the development of next-generation algorithms in job shop scheduling research.}



\maketitle


\section{Introduction}\label{sec:Intro}
Job shop scheduling problems (JSPs) are a fundamental category of scheduling challenges that are vital for optimizing operations in a wide range of industrial production environments \citep{zhang2019review, wang2021dynamic}. These problems are defined by a collection of jobs, each consisting of one or more operations that need to be scheduled on a set of machines. The inherent NP-hard nature of JSPs leads to significant complexity, which is often further amplified by additional context-specific constraints from real-world industrial environments. Given the diverse nature of production tasks and the variability between manufacturing sectors, numerous JSP variants have been introduced to address distinct scheduling requirements \citep{xiong2022survey}. Commonly observed variants and extensions of JSPs include the Flow Shop Problem (FSP), where operations across jobs must follow a fixed order on a series of machines; the Flexible Job Shop Problem (FJSP), which allows operations to be processed on multiple machines, adding a layer of routing flexibility; and the Assembly Job Shop Problem (AJSP), which incorporates constraints that require operations from different jobs to be coordinated for effective assembly processes. Another commonly observed variant is the inclusion of Sequence-Dependent Setup Times (SDST), which increases complexity by requiring set-up times that depend on the job sequence, a challenge often encountered in industries requiring frequent machine adjustments for specific job sequences. Each of these variants introduces unique challenges that further increase the complexity of finding optimal solutions.

Approaches for addressing JSP variants can be categorized into three categories: exact algorithms, heuristic algorithms, and learning-based methods. Exact algorithms guarantee optimal solutions but are computationally expensive and do not scale well for large problem instances. In contract, heuristic algorithms offer greater efficiency scalability, including improvement heuristics, such as evolutionary algorithms (e.g., \cite{zhang2011effective}), as well as construction heuristics (e.g., \cite{blackstone1982state, holthaus1997efficient}). However, heuristic methods often yield lower quality solutions and typically require extensive domain knowledge and significant tuning efforts. In recent years, learning-based approaches, particularly those that use deep reinforcement learning (DRL), have attracted considerable attention. DRL methods train neural networks to automatically learn effective heuristics, enabling efficient problem solving in a short computational time \citep{smit2024graph}.

Given the increasing complexity of industrial scheduling problems, which are gathering growing attention, and the rapid development of new solution methods, there is a clear need for more rigorous evaluation frameworks. Currently, researchers and practitioners face significant challenges in comparing different (learning and non-learning) approaches for JSP problem variants that incorporate practical constraints due to the scarcity of resources. The absence of a benchmark hinders thorough evaluations, restricts the use of consistent evaluation metrics, and prevents fair and reproducible comparisons of algorithms under standardized conditions. To address these challenges, we introduce a comprehensive benchmarking repository that covers a wide range of JSP variants along with their corresponding benchmark instances. This repository also provides a diverse set of solution methods, including both learning-based approaches and traditional techniques, all of which can be adapted to various scheduling scenarios. By providing a unified platform for evaluation, our resource enables researchers to benchmark new solution methods effectively while offering practitioners a solid foundation for initiating projects on industry-specific JSP variants or advancing algorithmic developments.

The remainder of this paper is organized as follows. Section 2 provides an overview of the different JSP variants. Section 3 details the technical implementation of the benchmark, describes the included solution methods, and demonstrates how to use the repository. Section 4 presents a comprehensive empirical evaluation that analyzes the performance of different solution methods across various JSP variants using publicly available benchmark instances. Lastly, Section 5 concludes the work.

\section{Job Shop Scheduling}
We have developed a scheduling environment capable of simulating the scheduling process for predefined problem instances and dynamic online environments where jobs arrive in real time. The current setup allows for scheduling across a variety of problem types, including JSP, FSP, FJSP, AJSP, variants with SDSTs, and variants where jobs arrive online according to specified distributions. In this section, we outline these problem variants using a disjunctive graph formulation: $G = (O, C, D)$, where $O = \{O_{ij} \mid \forall i, j\} \cup \{O_S, O_T\}$ represents the set of all operations. Here, $O_S$ and $O_T$ are dummy operations with zero processing time, serving as the start and end states \citep{blazewicz2000disjunctive}. $C$ represents the set of directed arcs (i.e., conjunctions), which denote precedence constraints between operations within the same job, while $D$ includes undirected arcs (i.e., disjunctions) linking operations that require the same machine for processing. Solving a scheduling problem instance involves determining the direction of each disjunction to ensure that the resultant graph forms a Directed Acyclic Graph (DAG). In the following, we outline each problem variant and provide its mathematical formulation.

\subsection{Problem Descriptions}
\begin{figure*}[h!]
    \centering
    \includegraphics[width=0.80\columnwidth, keepaspectratio]{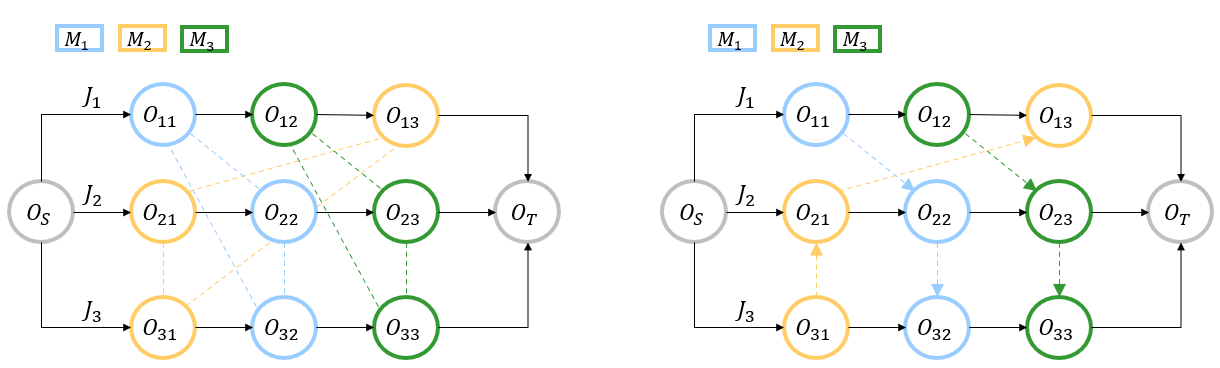}
    \caption{Disjunctive Graph Representation of a Job Shop Scheduling Problem (JSP). \emph{Left Panel}: This illustrates a 3 (jobs) $\times$ 3 (machines) JSP instance. Black arrows indicate conjunctive arcs, enforcing precedence constraints among operations within the same job. Dotted lines represent disjunctive arcs, whose directions need to be determined. Disjunctive arcs (or operation nodes) sharing the same color correspond to operations requiring the same machine. \emph{Right Panel}: Represents a feasible solution (best viewed in color).}
    \label{fig:jss}
\end{figure*}

In the \textit{Job Shop Scheduling Problem (JSP)},  \( n \) jobs \( J = \{J_1, J_2, \ldots, J_n\} \) must be scheduled across \( m \) machines \( M = \{M_1, M_2, \ldots, M_m\} \). Each job \( J_i \) consists of a series of operations \( O_{i,j} \), where $O_{i,j}$ represents the $j$-th operation of the $i$-th job. Each operation $O_{i,j}$ must be processed on a specific machine $M_k$ for a given processing time $p_{i,j,k}$. Operations must be scheduled such that no machine processes more than one operation at a time and operations within each job follow their specified order. The disjunctive graph formulation for JSP is shown in Figure \ref{fig:jss}, which illustrates a 3-job, 3-machine instance. Black arrows represent conjunctive arcs, denoting precedence within jobs, while dotted lines indicate disjunctive arcs, whose directions must be determined to create a valid schedule. Disjunctive arcs (or operation nodes) of the same color indicate operations that require the same machine.

\begin{figure*}[h!]
    \centering
    \includegraphics[width=0.5\columnwidth, keepaspectratio]{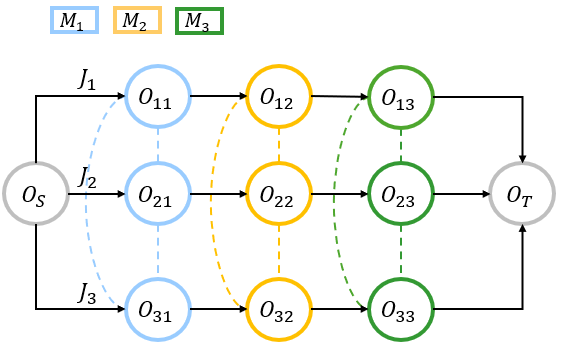}
    \caption{Disjunctive Graph Representation of a Flow Shop Scheduling Problem (FSP). This illustrates a 3 (jobs) $\times$ 3 (machines) FSP instance. All jobs follow the same fixed machine order, with same-colored operation nodes representing operations to be scheduled on the same machine.}
    \label{fig:fss}
\end{figure*}

\textit{Flow Shop Scheduling Problem (FSP)} differs from JSP by imposing a strict order across machines. In FSP, for any given job \( J_i \), operations must follow a fixed sequence across the machines: \( M_1 \rightarrow M_2 \rightarrow \ldots \rightarrow M_m \). Each operation \( O_{i,j} \) in job \( J_i \) corresponds to processing on a specific machine \( M_j \) and has a predetermined processing time \( p_{i,j} \). This restriction contrasts with the flexibility seen in JSP, where the order of operations can vary. The disjunctive graph formulation for FSP is illustrated in Figure \ref{fig:fss}, where the sequential nature of the problem is reflected in the structure of the graph.

\begin{figure*}[h!]
    \centering
    \includegraphics[width=0.5\columnwidth, keepaspectratio]{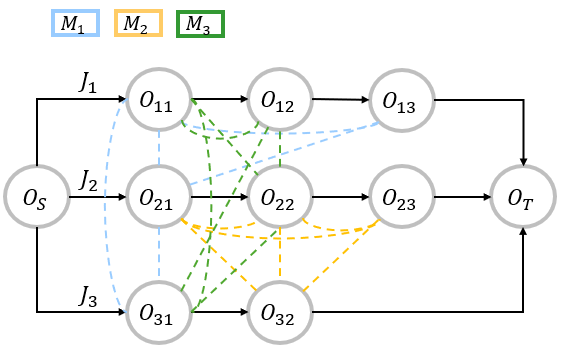}
    \caption{Disjunctive Graph Representation of a Flexible Job Shop Scheduling Problem (FJSP). This illustrates a 3 (jobs) $\times$ 3 (machines) FJSP instance. Operations, represented by gray nodes, can be assigned to different machines, allowing alternative routing options. Disjunctive arcs reflect both sequencing and machine assignment decisions, increasing scheduling complexity.}
    \label{fig:fjss}
\end{figure*}

In both the JSP and FSP problem variants, each operation \( O_{i,j} \) of a job \( J_i \) must be processed on a specific, predetermined machine \( M_k \) with a defined processing time. This differs in flexible problem variants, such as the \textit{Flexible Job Shop Scheduling Problem (FJSP)}, where each operation \( O_{i,j} \) of a job \( J_i \) can be processed on one or more machines. This flexibility introduces an extra layer of decision-making, requiring choices on how to allocate operations to machines, considering factors such as machine utilization and processing time to achieve the most efficient scheduling. The disjunctive graph formulation for FJSP is shown in Figure \ref{fig:fjss}.

\begin{figure*}[h!]
    \centering
    \includegraphics[width=0.65\columnwidth, keepaspectratio]{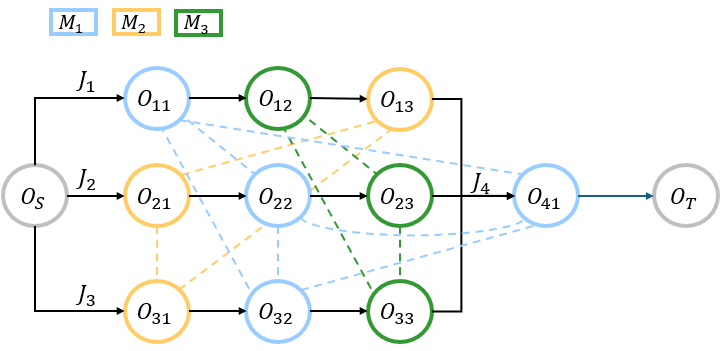}
    \caption{Disjunctive Graph Representation of an Assembly Job Shop Scheduling Problem (AJSP). This illustrates a 4 (jobs) $\times$ 3 (machines) AJSP instance. The first three jobs follow standard processing constraints, while the fourth job depends on their completion, enforcing an additional assembly precedence structure. Black arrows indicate the conjunctive arcs, representing both within-job precedence constraints and final assembly dependencies.}
    \label{fig:as}
\end{figure*}

In \textit{Assembly Job Shop Scheduling Problem (AJSP)}, a set of jobs is associated with different components that must be produced and assembled in a specific sequence to produce a final product. Each job corresponds to the assembly of a component, and the scheduling process is characterized by precedence constraints that dictate the order in which individual components must be manufactured prior to the assembly operation. This ensures that all necessary components are available and ready for assembly at the appropriate time. The disjunctive graph formulation for this scheduling variant is illustrated in Figure \ref{fig:as}.

In the \textit{Assembly Job Shop Scheduling Problem (AJSP)}, a set of jobs involves the production and assembly of different components in a specific sequence to create a final product. 

In the \textit{Assembly Job Shop Scheduling Problem (AJSP)}, a set of jobs is associated with various components that must be produced and assembled in a specific sequence to create a final product. The problem is defined by precedence constraints that dictate the order in which components must be processed on different machines before they can be assembled together. The assembly operations depend on the completion of multiple preceding operations before its start. The disjunctive graph formulation for this scheduling variant is shown in Figure \ref{fig:as}.

An additional layer of complexity is introduced when \textit{sequence-dependent setup times (SDSTs)} are considered. In this variant, setup times between operations on the same machine depend on the sequence in which the operations are scheduled. This introduces a new challenge, as both the processing times and the setup times between operations must be taken into account to minimize total processing time and improve the overall efficiency of the schedule. The disjunctive graph formulation for a scheduling problem with SDST is shown in Figure \ref{fig:sdst}, which illustrates the interaction between operations and their sequence-dependent setup requirements.

\begin{figure*}[h!]
    \centering
    \includegraphics[width=0.5\columnwidth, keepaspectratio]{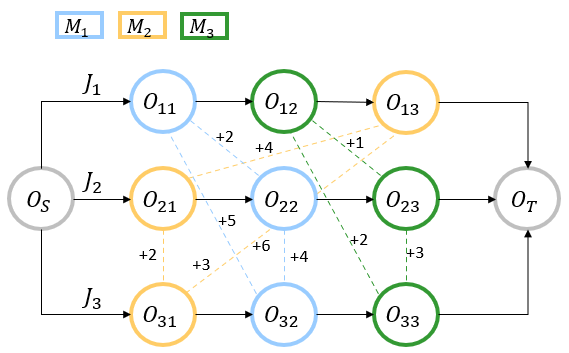}
    \caption{Disjunctive Graph Representation of a scheduling problem with Sequence-Dependent Setup Times (SDST). This illustrates a 3 (jobs) $\times$ 3 (machines) scheduling instance where disjunctive arcs not only determine operation sequencing on shared machines but also encode sequence-dependent setup times.}
    \label{fig:sdst}
\end{figure*} 

Beyond traditional static scheduling problems, \textit{dynamic scheduling} introduces an additional challenge by allowing jobs to arrive in real-time according to predefined probability distributions. Unlike static JSP and its variants, where all job details are known beforehand, dynamic scheduling requires adaptive scheduling algorithms capable of responding to changes in real-time. In the disjunctive graph representation, this means that the graph structure itself evolves over time: new operation nodes and their corresponding conjunctive arcs are incrementally added as jobs arrive. This dynamic nature demands adaptive scheduling strategies capable of making online decisions, ensuring that new operations are seamlessly accounted for in the scheduling.

A key feature of this scheduling environment is its ability to extend and combine multiple problem variants. Users can integrate additional constraints, modify objective functions, and construct hybrid scheduling scenarios that incorporate elements from different problem types. For example, it allows one to combine FJSP with SDSTs (FJSP-SDST) or introduce dynamic job arrivals into traditionally static problems. This flexibility makes the benchmark a powerful tool for researchers and practitioners to explore novel scheduling challenges, benchmark new solution methods, and develop advanced scheduling algorithms for diverse problem variants.

\section{Job Shop Scheduling Environment}\label{sec:jspenv-main}
\begin{figure}[h]
    \centering
    \includegraphics[width=0.59\textwidth]{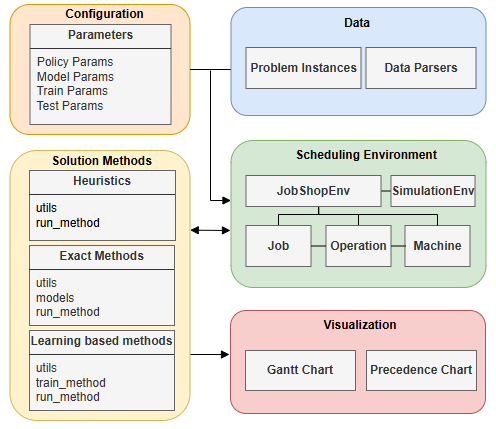}
    \caption{Graphical representation of the Job Shop Scheduling Benchmark repository. The benchmark consists of five main components: Configurations, Data, Scheduling Environment, Solution Methods, and Plotting. These components are described in Section \ref{sec:jspenv-main}, with each playing are role in configuring scheduling problems, interacting with solution methods, or visualizing results.}
    \label{fig:repo-figure}
\end{figure}

The proposed Job Shop Scheduling Benchmark is an open-source Python project hosted on GitHub. It is designed to provide a modular and extensible framework for solving and analyzing scheduling problems and solution methods. The repository is structured into five main components: configurations, data (problem instances), scheduling environment, solution methods, and visualization, as shown in Figure \ref{fig:repo-figure}. These components maintain essential properties for configuring scheduling problems and provide a unified interface to interact with solution methods to generate and evaluate scheduling solutions. We will discuss these components in the following subsections. 

\subsection{Scheduling Environment}\label{sec:jspenv}
The core of the Job Shop Scheduling Benchmark is its scheduling environment, consisting of four essential classes that form its foundation: \textit{Job}, \textit{Operation}, \textit{Machine}, and \textit{JobShopEnv}. The \textit{Job} class captures the key attributes of individual jobs. Each job in the system is identified by a unique job identifier and consists of a list of associated operations that need to be scheduled. The \textit{Operation} class contains detailed information about specific operations within a job, including its unique Operation identifier, the corresponding Job identifier, and processing times for different machine alternatives. The \textit{Machine} class represents machine resources within the scheduling environment. Each machine is assigned a unique machine identifier and keeps a list to track the operations assigned to it, along with their scheduling details (i.e., start times, process times, and setup times). The \textit{JobShopEnv} class serves as the central hub for managing all jobs, operations, and machine interactions. It integrates the configured Job, Operation, and Machine classes into a unified environment and facilitates the scheduling processes. This class provides functionalities for the tracking and scheduling of operations throughout the scheduling process. With this, the JobShopEnv acts as the interface through which solution methods can interact with the scheduling problem, ensuring that scheduling methods have access to a standardized setup for various configured jobs shop scheduling problem variants.

\subsection{Static and Dynamic Scheduling}
The environment supports both static and dynamic scheduling scenarios through two key components: \textit{Data Parsers} and a \textit{SimulationEnv}. For static problem instances, we developed several parsers capable of handling various problem variants and data formats commonly found in the literature. These parsers extract information from instance files and configure the \textit{JobShopEnv} with the corresponding \textit{Job}, \textit{Operation}, and \textit{Machine} objects. This ensures that solution methods can operate within a standardized environment in which different variants of problems can be initialized from problem instances from the literature. 
We also offer a JSON-based data format with an accompanying parser, allowing practitioners to create custom problem instances without relying on the often cumbersome data formats commonly used in the literature (see Section \ref{sec:example_json}). The \textit{SimulationEnv} class provides a discrete event simulation environment tailored for real-time job arrivals. It manages job arrival details within the JobShop and generates the corresponding Job and Operation objects based on a specified distribution. This class is optional and is only required for online environments where jobs arrive dynamically according to a pre-defined distribution.

\subsection{Configuration and Visualization}
The interaction within the repository is configured through \textit{Configuration} files. These files define all the required parameters for configuring the solution methods for testing and training (i.e., for learning-based methods), making it easy to test different scheduling scenarios and solution methods without needing to modify the underlying code. Problem instances and the scheduling solutions obtained can be inspected through \textit{Precedence Charts} and \textit{Gantt charts}. The precedence charts show the order in which operations must be performed within each job, while the Gantt charts display the timeline of operations scheduled across machines, helping users evaluate the scheduling efficiency of found solutions.

\begin{figure*}[h!]
    \centering
    \begin{subfigure}[b]{0.75\columnwidth}
        \centering
        \includegraphics[width=\linewidth, keepaspectratio]{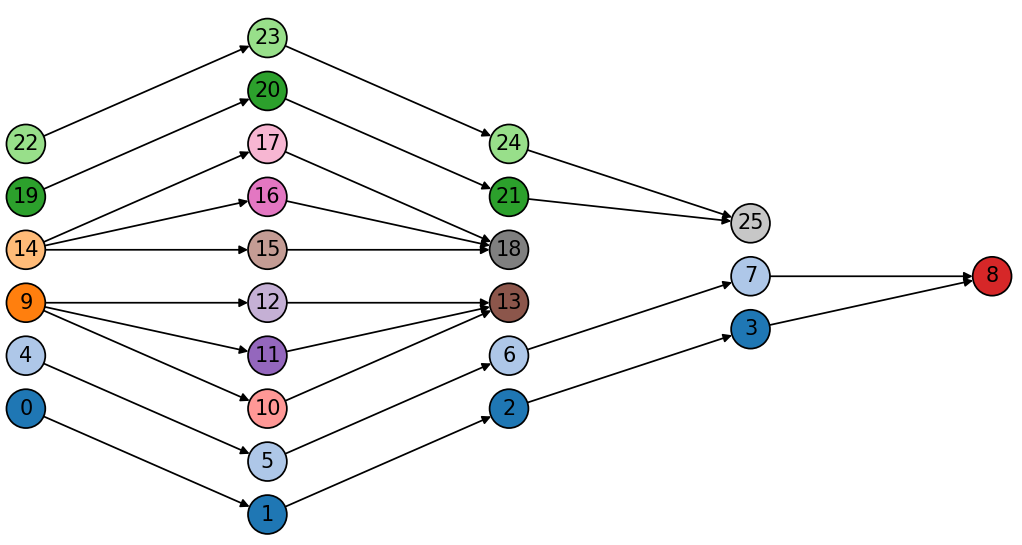}
        \caption{ }
        \label{fig:convergence}
    \end{subfigure}
    
    \vspace{1cm} 

    \begin{subfigure}[b]{0.85\columnwidth}
        \centering
        \includegraphics[width=\linewidth, keepaspectratio]{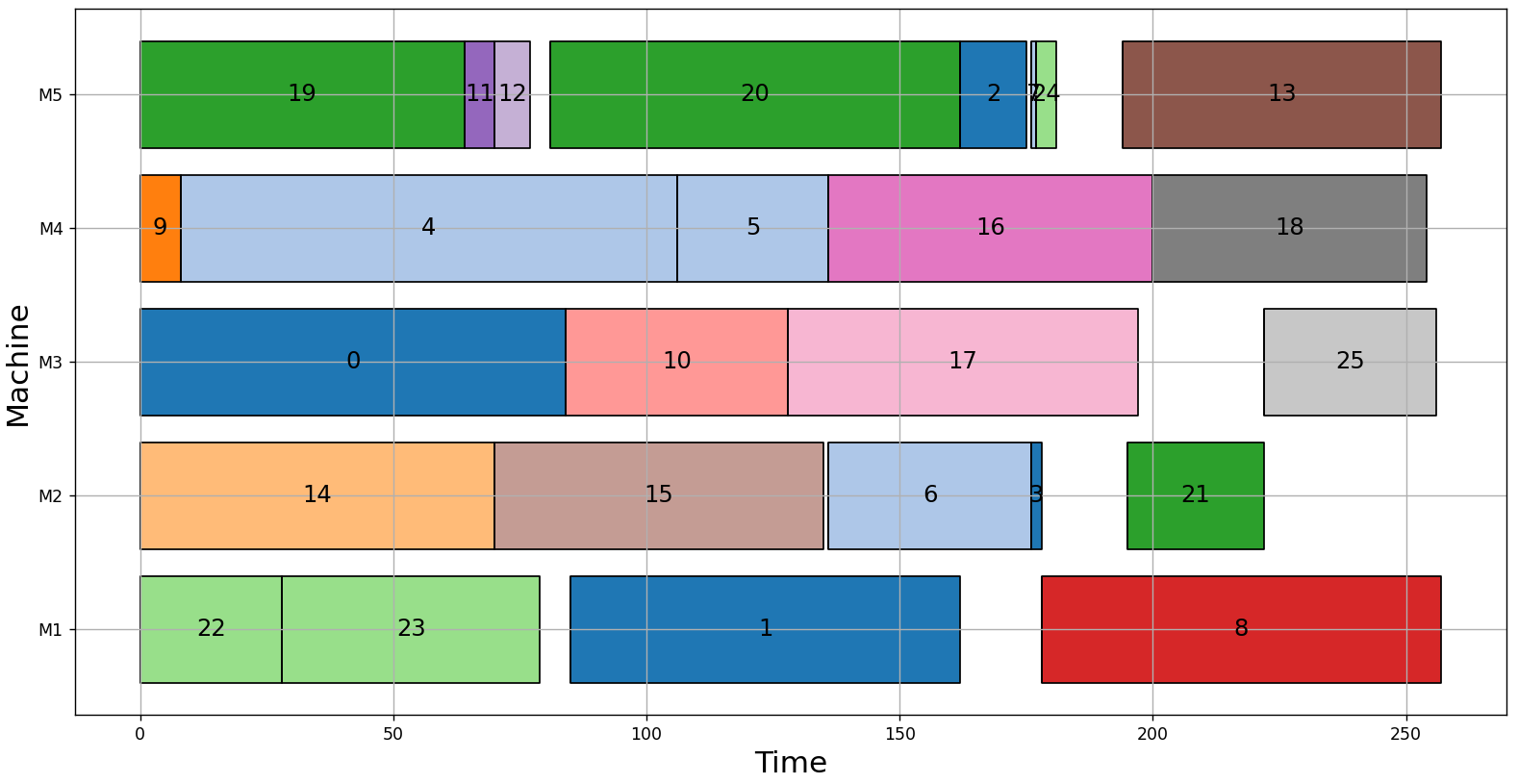}
        \caption{ }
        \label{fig:sdst}
    \end{subfigure}
    
    \caption{(a) Precedence Chart of an instance with assembly constraints, where operations wait for more than one predecessor from different jobs to be completed. Colors represent the job ID, and each operation is assigned a unique operation ID. (b) Gantt Chart of a solution for the same instance, illustrating the the machine allocation and sequence of scheduled operations.}
    \label{fig:combined}
\end{figure*}

\subsection{Solution Methods}
The Job Shop Scheduling Benchmark environment provides a diverse range of solution methods tailored to address the scheduling problems configured within the environment. These methods can be broadly classified into three categories: exact algorithms, heuristic algorithms, and learning-based methods. Each solution method is implemented to seamlessly interact with the \textit{JobShopEnv} introduced in Section \ref{sec:jspenv} for solving a problem variant, and Table \ref{tab:solution_methods} summarizes their applicability to various problem variants. The included solution methods are described in detail below:

\begin{table}[htb!]
    \centering
    \small
    \renewcommand{\arraystretch}{1.3}
    \begin{tabular}{l|cccccc}
        \hline
        \textbf{Solutions methods} & \textbf{JSP} & \textbf{FSP} & \textbf{FJSP} & \textbf{SDST} & \textbf{AJSP} &\textbf{Dynamic JSP} \\
        \hline
        Exact: CP-SAT & \cmark & \cmark & \cmark & \cmark & \cmark & \xmark \\
        Exact: MILP & \cmark & \cmark & \cmark & \cmark & \cmark & \xmark \\
        Heuristics: Priority Dispatching Rules & \cmark & \cmark & \cmark & \cmark & \cmark & \cmark \\
        Heuristics: Genetic Algorithm & \cmark & \cmark & \cmark & \cmark & \cmark & \xmark \\
        DRL: L2D & \cmark & \cmark & \xmark & \xmark & \xmark & \\
        DRL: FJSP-DRL & \cmark & \cmark & \cmark & \xmark & \xmark & \cmark \\
        DRL: DANIEL & \cmark & \cmark & \cmark & \xmark & \xmark & \\
        \hline
    \end{tabular}
    \caption{Available implementation of different solution methods across different scheduling problem variants in the benchmark environment, as of February 6th, 2025. A check mark (\cmark) indicates that the solution method is implemented and capable of solving the given scheduling problem variant, while a cross mark (\xmark) indicates that the method is not capable of solving it. The repository continues to be updated with new solution methods and instances.}    
    \label{tab:solution_methods}
\end{table}

\textbf{Exact algorithms} are designed to find optimal solutions by exhaustively exploring the solution space, ensuring the identification of optimal job sequences and machine assignments for job shop scheduling problems. These approaches, such as branch-and-bound, dynamic programming, and integer programming, are guaranteed to identify the best possible solutions, when provided with sufficient computing resources. However, as the size of the problem grows, these methods can become inefficient and face scalability challenges, making them most appropriate for small to medium-sized problems where computational capacity is sufficient. In the Job Shop Scheduling Benchmark, two primary exact approaches are implemented: Mixed Integer Linear Programming (MILP) and Constraint Programming (CP).

\textit{Mixed Integer Linear Programming (MILP)} is a widely used method for solving constraint optimization problems (COPs) due to its flexibility in handling a wide range of constraints. By combining continuous and integer decision variables within linear objectives and constraints, MILP offers a versatile and efficient approach to addressing a wide range of optimization challenges. The repository integrates the MILP models developed by \citet{manne1960job}, \citet{roshanaei2013mathematical}, and \citet{shen2018solving} to address various scheduling problems with different constraints. These MILP models are defined to work with the Gurobi optimizer, which employs techniques such as branch-and-bound, cutting planes, and branch-and-cut to explore the solution space.


\textit{Constraint Programming (CP)} is another exact method used in the environment, particularly well suited to handle complex combinatorial problems. In the context of job shop scheduling, CP is effective at modeling intricate constraints such as precedence relations, machine-specific requirements, and resource allocations. It uses techniques such as constraint propagation, backtracking, and domain reduction to explore possible solutions. CP solvers, such as IBM ILOG CPLEX and OR-Tools, are highly flexible and capable of addressing a wide range of constraints, making CP a powerful tool for solving diverse scheduling problems. In context of this benchmark, we specifically use the CP-SAT solver from OR-Tools, which is known for its efficiency in solving complex scheduling problems. More specifically, we have extended the CP formulations provided in \cite{perron2023} to fit the specifics of the job shop scheduling problem variants central in our proposed benchmarking repository.

\textbf{Heuristic algorithms} are designed to find good enough solutions in a reasonable time frame, sacrificing optimality for efficiency. Such methods, which include dispatching rules, genetic algorithms, and various local search strategies, are generally more efficient and scalable than exact algorithms, making them more suitable for large-scale problems. However, the quality of the solutions they find can vary and is heavily dependent on the domain knowledge used to design and tune the heuristics.

\textit{Priority Dispatching Rules (PDRs)} are a class of heuristic algorithms that determine the order of job processing by prioritizing them according to specific criteria, such as job arrival times, operation durations or resource requirements. In the Job Shop Scheduling Benchmark, we implement several commonly used PDRs, including First In, First Out (FIFO), Least Operations Remaining (LOR), Least Work Remaining (LWR), Most Operations Remaining (MOR), and Most Work Remaining (MWR). FIFO, being one of the simplest and most intuitive, processes jobs in the order of their arrival. LOR and LWR prioritize jobs that have the least remaining processing, facilitating their early completion. In contrast, MOR and MWR focus on balancing the overall completion times by prioritizing jobs with the most remaining operations and work, respectively. In addition to these PDRs, machine allocation rules are used in Flexible problem variants (e.g., FJSP), for allocating operations to machines. Two commonly used rules are Earliest Expected Time (EET) and Shortest Processing Time (SPT). EET assigns jobs to the machine that is expected to become available soonest, reducing machine idle time and enhancing overall throughput. SPT prioritizes jobs with the shortest processing time.

\textit{Genetic Algorithms (GAs)} are search-based heuristics inspired by the principles of natural selection. They iteratively evolve solutions by simulating biological processes, such as selection, crossover, and mutation, to identify optimal or near-optimal solutions. Although GAs are generally slower than PDRs due to their computationally intensive search process, their evolutionary approach is highly effective at uncovering innovative and diverse solutions that simpler methods might overlook. In the Job Shop Scheduling Benchmark, we specifically implement a GA configured with two main components: a Machine Selection component and an Operation Sequence component, based on \cite{zhang2011effective}. The Machine Selection component is an integer array that assigns operations to specific machines. Each operation has a fixed position, and the value at that position corresponds to the selected machine for the operation. The Operation Sequence component dictates the scheduling order of operations on assigned machines, allowing each job index to appear as many times as its number of operations. This encoding enables the construction of schedules by assigning operations to the earliest feasible times on their allocated machines. The GA population is initialized with three methods: Global Selection (minimizing total machine processing time across all jobs), Local Selection (minimizing machine time for individual jobs), and Random Selection. The Operation Sequence is initialized randomly. The Machine Selection component undergoes two-point or uniform crossover, while the Operation Sequence component uses precedence-preserving order-based crossover (POX), which maintains the scheduling order of a subset of jobs. This approach effectively balances solution diversity and optimization by initializing 60\% of the population with Global Selection, 30\% with Local Selection, and 10\% randomly, enhancing its ability to find near-optimal schedules.

\textbf{Learning-based methods} have gained significant attention in solving job shop scheduling problems. Among these methods, DRL and graph neural networks (GNNs) have proven to be particularly effective. These approaches are well-suited for handling scheduling problems because they can automatically learn effective heuristics and policies by interacting with the problem environment. Since scheduling problems can naturally be represented as graphs, where jobs and machines correspond to nodes and edges, GNNs are ideal for capturing the relationships among jobs, machines, and their constraints. GNNs are capable of learning the structure of the scheduling problem and generating efficient heuristics that can tackle various problem configurations  ~\citep{xie2019review,xiong2022survey, smit2024graph}. In the context of this benchmark, we have included several representative DRL-based methods as follows.

\textit{Learning-to-Dispatch (L2D)}, proposed by \citet{zhang2020learning}, introduced the first end-to-end DRL approach to address JSP. L2D learns PDRs using a disjunctive graph representation, with a graph isomorphism network (GIN) trained via proximal policy optimization (PPO) to generate embeddings for the graph, which are then used to make scheduling decisions. L2D consistently outperformed traditional PDRs in terms of solution quality while maintaining high speed. Although L2D does not achieve the performance of exact solution methods, such as the OR-Tools CP-SAT solver, it provides significantly faster processing times on larger instances.

\textit{FJSP-DRL}, introduced by \cite{song2022flexible}, extends L2D to learn PDRs to solve FJSP. This method extends the disjunctive graph by introducing machine nodes, creating a heterogeneous graph structure that effectively captures the relationships among operations, machines, and their connections. This reduces graph density and allows the trained DRL agent to consider the full scope of scheduling constraints by combining operation selection and machine assignment decision making into a single decision. A heterogeneous graph neural network (HGNN) is used to generate feature embeddings for both operations and machines, which are input to a policy network trained via the PPO algorithm. Experiments demonstrate that this approach outperforms traditional PDRs by achieving higher scheduling quality and maintaining computational efficiency on larger problem instances. Besides its application to FJSP, the approach can still be applied to less complex variants, such as Job Shop Scheduling and Flow Show Scheduling without the machine selection problem characteristic. 

\textit{DANIEL}, proposed by \cite{wang2023flexible} makes use of self-attention models and DRL to solve FJSP. The framework introduces a dual attention network (DAN) to map the priority of operations and the relationship of machine selection through separate attention mechanisms. This allows for the processing of large and complex schedules while reducing irrelevant data, making it a powerful approach for larger-scale scheduling problems. Trained using PPO, the framework demonstrates strong performance, outperforming traditional PDRs and other DRL-based methods.

\subsection{Example usage}
Building on the solution methods discussed in the previous section, we present 3 example usages of different solution methods to interact with the environment to obtain solutions. The first example in Listing 1 shows how to use the PDR methods. The different parameters for configuring the problem instance to be solved and the configuration of the solution method (its dispatching rules) are provided through a configuration file. The second example in Listing 2 shows how a GA can be used to solve a problem configured in the environment, and the third example in Listing 3 shows how a DRL-based method, L2D, can be used. These examples provide a helpful understanding of how the different solution methods interact with the environment. We provide more extensive implementation examples for each of the different solution methods alongside their implementation code in the repository\footnote{\url{https://github.com/ai-for-decision-making-tue/Job_Shop_Scheduling_Benchmark_Environments_and_Instances/blob/main/tutorial_benchmark_environment.ipynb}}\footnote{\url{https://github.com/ai-for-decision-making-tue/Job_Shop_Scheduling_Benchmark_Environments_and_Instances/blob/main/tutorial_custom_problem_instance.ipynb}}

\begin{lstlisting}[style=pythonstyle, caption={Dispatching Rules example}]
from solution_methods.dispatching_rules import run_dispatching_rules
from solution_methods.helper_functions import load_job_shop_env, load_parameters

parameters = load_parameters("configs/dispatching_rules.toml")
jobShopEnv = load_job_shop_env(parameters['instance'].get('problem_instance'))

makespan, jobShopEnv = run_dispatching_rules(jobShopEnv, **parameters)
\end{lstlisting}

\begin{lstlisting}[style=pythonstyle, caption={Genetic Algorithm example}]
from solution_methods.GA.src.initialization import initialize_run
from solution_methods.helper_functions import load_job_shop_env, load_parameters
from solution_methods.GA.run_GA import run_GA

parameters = load_parameters("configs/genetic_algorithm.toml")
jobShopEnv = load_job_shop_env(parameters['instance'].get('problem_instance'))

population, toolbox, stats, hof = initialize_run(jobShopEnv, **parameters)
makespan, jobShopEnv = run_GA(jobShopEnv, population, toolbox, stats, hof, **parameters)  
\end{lstlisting}

\begin{lstlisting}[style=pythonstyle, caption={L2D example}]
from solution_methods.L2D.src.run_L2D import run_L2D
from solution_methods.helper_functions import load_job_shop_env, load_parameters
   
parameters = load_parameters("configs/L2D.toml")
jobShopEnv = load_job_shop_env(parameters['instance'].get('problem_instance'))
makespan, jobShopEnv = run_L2D(jobShopEnv, **parameters)
\end{lstlisting}

\subsection{Custom problem configuration}\label{sec:example_json}
In addition to supporting standard problem instances from the literature, the Job Shop Scheduling Environment allows users to define custom scheduling problems using a flexible JSON-based format. This format enables users to specify the elements of their problem configuration, such as scheduling type (job shop or flow shop), machine assignments, job dependencies, operation-specific processing times, and sequence-dependent setup times. Through this, users can tailor problem configurations to suit specific operational needs or research scenarios and pass a configured environment to a solution method. An example of this JSON configuration is provided in Listing 4.

\begin{lstlisting}[style=pythonstyle, caption={Custom problem example}]
processing_info = {
    "instance_name": "custom_problem_instance",
    "nr_machines": 2, 
    "jobs": [
        {"job_id": 0, "operations": [
            {"operation_id": 0, "processing_times": {"machine_1": 10, "machine_2": 20},
            "predecessors": None},
            {"operation_id": 1, "processing_times": {"machine_1": 25, "machine_2": 19},
            "predecessors": [0]}
        ]},
        {"job_id": 1, "operations": [
            {"operation_id": 2, "processing_times": {"machine_1": 23, "machine_2": 21},
            "predecessors": None},
            {"operation_id": 3, "processing_times": {"machine_1": 12, "machine_2": 24},
            "predecessors": [2]}
        ]},
        {"job_id": 2, "operations": [
            {"operation_id": 4, "processing_times": {"machine_1": 37, "machine_2": 21},
            "predecessors": None},
            {"operation_id": 5, "processing_times": {"machine_1": 23, "machine_2": 34},
            "predecessors": [4]}
        ]}
    ],
    "sequence_dependent_setup_times": {
        "machine_1": [
            [0, 25, 30, 35, 40, 45],
            [25, 0, 20, 30, 40, 50],
            [30, 20, 0, 10, 15, 25],
            [35, 30, 10, 0, 5, 10],
            [40, 40, 15, 5, 0, 20],
            [45, 50, 25, 10, 20, 0]
        ],
        "machine_2": [
            [0, 21, 30, 35, 40, 45],
            [21, 0, 10, 25, 30, 40],
            [30, 10, 0, 5, 15, 25],
            [35, 25, 5, 0, 10, 20],
            [40, 30, 15, 10, 0, 25],
            [45, 40, 25, 20, 25, 0]
        ]
    }
}
\end{lstlisting}

Consequently, in Listing 5, we show how the GA can be applied to solve the custom-configured problem. The listing illustrates how the problem setup is integrated into the algorithm, which then iteratively refines the scheduling solutions through crossover, mutation, and selection.

\begin{lstlisting}[style=pythonstyle, caption={Example for solving the custom configured problem with a Genetic Algorithm}]
from solution_methods.GA.src.initialization import initialize_run
from solution_methods.GA.run_GA import run_GA
from data.data_parsers.custom_instance_parser import parse

parameters = load_parameters("configs/genetic_algorithm.toml")
jobShopEnv = parse(processing_info)
population, toolbox, stats, hof = initialize_run(jobShopEnv, **parameters)
makespan, jobShopEnv = run_GA(jobShopEnv, population, toolbox, stats, hof, **parameters)

\end{lstlisting}

\section{Experimentation}
We present a comprehensive empirical evaluation that analyzes the performance of different solution methods across various variants of Job Shop Scheduling problems, using publicly available benchmark instances for an in-depth comparison. In this section, we first introduce the available benchmarks, followed by the experimental setup, and then provide the performance results from our experiments on a selected set of publicly available problem instances.

\subsection{Problem Instances}
Over the years, various works have introduced problem instances for different variants of scheduling problems. Many of these have become widely accepted benchmarks in subsequent research. These standardized datasets facilitate fair comparisons between algorithms, offering a consistent framework for evaluating performance across a wide array of problem configurations. In the context of this project, we have collected various datasets and provided them in our repository. We provide an overview in Table \ref{tab:instances} of the included datasets. The table provides an overview of five major scheduling problem categories that can be modeled and solved using the environment: JSP, FSP, FJSP, FJSP with SDSTs (FJSP-SDST), and a Flexible AJSP (FAJSP). For each category, the table details the original studies or datasets from which these benchmarks were derived, the problem scale (e.g., number of jobs and machines), and the number of instances.

\begin{table}[htb!]
\centering
\footnotesize
\renewcommand{\arraystretch}{1.3}
\setlength{\tabcolsep}{10pt}
\caption{Public benchmarking instances for scheduling problems.}
\begin{tabular}{p{4.4cm} | p{10.7cm}}
\hline
\textbf{Scheduling Problems} & \textbf{Problem Benchmarks and Details} \\ \hline
Job Shop Scheduling (JSP) & 
    \vspace{-0.5em} 
    \begin{itemize}\setlength\itemsep{0em}\setlength\parsep{0em}
        \item \cite{adams1988shifting}: 10-20 jobs, 5-10 machines, 6 instances.
        \item \cite{demirkol1998benchmarks}: Medium-scale, 20 instances.
        \item  \cite{taillard1993benchmarks}: Up to 100 jobs and 20 machines, 80 instances.
    \end{itemize}
    \vspace{-0.5em} \\ \hline 
Flow Shop Scheduling (FSP) & 
    \vspace{-0.5em}
    \begin{itemize}\setlength\itemsep{0em}\setlength\parsep{0em}
        \item \cite{vallada2015new}: 10-800 jobs, 5-60 machines, 2x240 instances.
    \end{itemize}
    \vspace{-0.5em} \\ \hline
Flexible Job Shop Scheduling (FJSP) & 
    \vspace{-0.5em}
\begin{itemize}\setlength\itemsep{0em}\setlength\parsep{0em}
        \item \cite{brandimarte1993routing}: 10-20 jobs, 4-15 machines, 10 instances.
        \item \cite{hurink1994tabu}: 7-30 jobs, 4-15 machines, 4x66 instances.
        \item \cite{dauzere1994solving}: 10-20 jobs, 5-10 machines, 18 instances.
        \item \cite{barnes1996flexible}: 10-15 jobs, 11-18 machines, 21 instances.
        \item \cite{kacem2002pareto}: High flexibility, 8 instances.
        \item \cite{fattahi2007mathematical}: 2-12 jobs, 2-8 machines, 20 instances.
    \end{itemize}
    \vspace{-0.5em} \\ \hline
FJSP with Sequence Dependent Setup Times (FJSP-SDST) & 
    \vspace{-0.5em}
    \begin{itemize}\setlength\itemsep{0em}\setlength\parsep{0em}
        \item \cite{fattahi2007mathematical} (extended by Hexaly): 2-12 jobs, 2-8 machines, 20 instances.
    \end{itemize}
    \vspace{-0.5em} \\ \hline
Flexible Assembly Job Shop Scheduling (FAJSP) & 
    \vspace{-0.5em}
    \begin{itemize}\setlength\itemsep{0em}\setlength\parsep{0em}
        \item \cite{birgin2014milp}: 10-50 jobs, 5-10 machines, 50 instances.
    \end{itemize}
    \vspace{-0.5em} \\ \hline
\end{tabular}
\label{tab:instances}
\end{table}

\subsection{Experimental Setup}
We conducted experiments on JSP, FSP, and FJSP with the benchmarking instances provided above to demonstrate the performance of different algorithms within the benchmarking environment. For each type of problem, we selected specific subsets of instances. For JSP, we used the first 80 instances from \cite{taillard1993benchmarks}. For FSP, the experiments were based on instances from \cite{vallada2015new}, with 10, 30, and 50 jobs. For FJSP, we used instances from \cite{brandimarte1993routing} as well as instances e, r, and vdata from \cite{hurink1994tabu}.

The experiments were executed on a system equipped with an Intel Core i7 processor, 32GB of RAM. For exact methods, we used CP-SAT with the OR-Tools solver (v9.9) and MILP with Gurobi (v11.0.1). Both solvers were configured with a maximum search time of 3600 seconds for each instance. For heuristic methods, we applied the SPT, MWR, and MOR dispatching rules. In case of the FJSP problem variant, an earliest-available heuristic was used for machine selection, ensuring that jobs were assigned to the machine that would be available the earliest. These heuristic methods are computationally lightweight and do not require additional parameters to be set. A GA was configured with a population size of 100 and a search budget of 100 generations. The algorithm was configured with a crossover rate of 0.7, and the mutation rate was set at 0.02, following commonly used values for GA. For learning-based methods, we used L2D, FJSP-DRL, and DANIEL DRL-based methods. All three methods were trained on the same set of problem instances: 5x5, 10x10, and 20x20, representing the number of jobs and machines, respectively. For FJSP-DRL and DANIEL, both trained on the FJSP problem, we followed the instance generation configuration of \cite{song2022flexible}. The training hyperparameters were set according to those specified in their original publications. For learning-based solutions, we used greedy action selection to choose the most valued actions and an action sampling strategy to generate 100 distinct solutions per instance by sampling from the action values.

\subsection{Results on public instances}
We compare the obtained results with the best-known solutions (BKSs) from the literature and show the aggregated results in Table \ref{tab:agg_results}. We provide a full overview of the best-known solutions in Appendix \ref{appendix:best_known_solutions} and more detailed results tables in Appendix \ref{appendix:extended_results}.

\begin{table}[h!]
\centering
\caption{Aggregated Performance results for JSP, FSP and FJSP instances.}
\small
\begin{tabular}{lcccccccc}
\hline
\multicolumn{1}{c}{} & \multicolumn{2}{c}{JSP} &  & \multicolumn{2}{c}{FSP} &  & \multicolumn{2}{c}{FJSP} \\ \cline{2-3} \cline{5-6} \cline{8-9} 
\multicolumn{1}{c}{} & Mean makespan    & Mean gap    &  & Mean makespan    & Mean gap    &  & Mean makespan     & Mean gap    \\ \hline
CP-SAT               & 2380,56    & 0,75       &  & 2448,15    & 2,72       &  & 1433,67     & 0,09       \\
MILP                 & -          & -          &  & -          & -          &  & 1453,90     & 1,65       \\
SPT                  & 2951,98    & 27,51      &  & 2749,92    & 16,48      &  & 1754,51     & 23,88      \\
MWR                  & 2772,06    & 19,53      &  & 2968,63    & 24,78      &  & 1656,40     & 14,83      \\
MOR                  & 2778,00    & 19,69      &  & 2954,90    & 24,54      &  & 1656,42     & 14,83      \\
GA - 100x100         & 3101,26    & 34,50      &  & 2740,13    & 14,32      &  & 1765,70     & 23,70      \\
L2D - greedy         & 2930,24    & 26,58      &  & 2998,55    & 26,37      &  & -           & -          \\
L2D - sample         & 3072,71    & 30,92      &  & 3580,13    & 44,81      &  & -           & -          \\
FJSP-DRL - greedy    & 2898,20    & 23,93      &  & 2947,27    & 24,03      &  & 1721,10     & 20,41      \\
FJSP-DRL - sample    & 2767,16    & 18,47      &  & 2748,82    & 15,33      &  & 1555,40     & 8,61       \\
DANIEL - greedy      & 2740,28    & 18,24      &  & 2846,55    & 20,73      &  & 1658,03     & 14,94      \\
DANIEL - sample      & 2661,26    & 14,42      &  & 2750,73    & 16,71      &  & 1572,74     & 9,11       \\ \hline
BKS                  & 2366,65    & 0,00       &  & 2360,95    & 0,00       &  & 1432,80     & 0,00       \\ \hline
\end{tabular}
\label{tab:agg_results}
\end{table}

The CP-SAT solver stands out as the most effective method overall, consistently achieving the smallest mean gaps across all three types of problems. For JSP, it achieved a mean gap of only 0.75\%, closely approximating the Best-Known Solutions (BKS). For FSP, configured with larger problem instances, its achieves a mean gap of 2.72\%, and for FJSP, the solver achieves a mean gap of 0.09\%. These results highlight the solver's reliability and effectiveness in finding near-optimal solutions across diverse scheduling problems and sizes. However, it is important to note that this performance superiority comes with a significant computational cost, requiring much longer runtimes compared to heuristics and learning-based approaches. In contrast, the MILP method, being configured with the same runtime budget as CP-SAT, performed substantially worse. It was unable to produce solutions for many JSP and FSP problem instances within the configured time limits. 

The priority dispatching rules, such as SPT, MWR, and MOR provided quick solution to the included problem instances, but demonstrate significant performance gaps compared to CP-SAT and learning-based methods. For JSP, these heuristics achieved mean gaps ranging from 19.53\% (MWR) to 27.51\% (SPT), compared to BKS. In FSP, the gaps were slightly smaller, with SPT performing the best among the heuristics at 16.48\%, and in FJSP, MWR and MOR performed better, both achieving a mean gap of 14.83\%, while SPT fell behind with a gap of 23.88\%. This highlights the variability of heuristics depending on the type of problem. The GA exhibited inconsistent performance between problem types. In JSP, it performed poorly, achieving the largest mean gap of 34.50\%. However, in FSP, GA performance improved, with a mean gap of 14.32\%, making it competitive with learning-based methods. Overall, GA performance suggests that it requires a higher search budget (i.e., larger population size and/or more generations of search). In addition, it could also benefit from tuning its operator parameters (e.g., the work of \cite{zhang2011effective} notices the wide range of optimal parameters for different distributed problem instances of different sizes) or from being configured with better initial solutions (e.g., those obtained using the PRDs or obtained with the learning-based methods).

The learning-based methods, L2D, FJSP-DRL, and DANIEL, demonstrated competitive performance, often surpassing traditional heuristics. Among these, FJSP-DRL (sampling-based) was particularly strong, achieving the smallest mean gap for FJSP with an  8.61\%, excluding the exact methods, and also performed competitively for JSP and FSP. Also, the sampling configuration of DANIEL performed strongly, achieving a mean gap of 14.42\% for JSP, outperforming all heuristics and learning-based methods, and second best in FSP and FJSP (excluding the exact methods). The performance of the L2D method lags behind that of the other learning-based methods.

\section{Conclusion}
This benchmark repository is an ongoing and joint effort to address machine scheduling problems. New scheduling solutions using a hybrid approach, i.e., integrated learning and heuristics,  
are also in development and will be included upon completion. This repository also has welcomed contributions from researchers outside of the core development team, and will continue to be updated over time to provide a comprehensive profile of state-of-the-art performance in solving a wide range of machine scheduling problems.

\newpage
\begin{APPENDICES}
\section{Best known solutions}\label{appendix:best_known_solutions}
For the problem instances examined in the experiments above, the best known solutions were collected from three sources \footnote{The best known solutions for JSP are available at \url{http://optimizizer.com/TA.php}, for FSP instances see \url{https://ars.els-cdn.com/content/image/1-s2.0-S0377221714005992-mmc1.pdf}, and for FJSP, refer to \cite{schutt2013scheduling}.}

\begin{table}[H]
\footnotesize
\caption{Best-Known solutions (BKS) for \citet{taillard1993benchmarks} Job Shop Scheduling instances 1-80. Values in \textbf{bold} indicate optimal solutions.}
\centering
\renewcommand{\arraystretch}{0.9}
\begin{tabular}{ccccccccccc}
\hline
\multicolumn{2}{c}{Tailard 1 - 20} &  & \multicolumn{2}{c}{Tailard 21 -   40} &  & \multicolumn{2}{c}{Tailard 41 -   60} &  & \multicolumn{2}{c}{Tailard 61 -   80} \\ \cline{1-2} \cline{4-5} \cline{7-8} \cline{10-11} 
Instance & \multicolumn{1}{c}{BKS} &  & Instance   & \multicolumn{1}{c}{BKS}  &  & Instance   & \multicolumn{1}{c}{BKS}  &  & Instance   & \multicolumn{1}{c}{BKS}  \\ \hline
ta01     & \textbf{1231}           &  & ta21       & \textbf{1642}            &  & ta41       & 2006                     &  & ta61       & \textbf{2868}            \\
ta02     & \textbf{1244}           &  & ta22       & 1600                     &  & ta42       & 1939                     &  & ta62       & \textbf{2869}            \\
ta03     & \textbf{1218}           &  & ta23       & 1557                     &  & ta43       & 1846                     &  & ta63       & \textbf{2755}            \\
ta04     & \textbf{1175}           &  & ta24       & \textbf{1644}            &  & ta44       & 1979                     &  & ta64       & \textbf{2702}            \\
ta05     & \textbf{1224}           &  & ta25       & 1595                     &  & ta45       & 2000                     &  & ta65       & \textbf{2725}            \\
ta06     & \textbf{1238}           &  & ta26       & 1643                     &  & ta46       & 2006                     &  & ta66       & \textbf{2845}            \\
ta07     & \textbf{1227}           &  & ta27       & 1680                     &  & ta47       & 1889                     &  & ta67       & \textbf{2825}            \\
ta08     & \textbf{1217}           &  & ta28       & \textbf{1603}            &  & ta48       & 1937                     &  & ta68       & \textbf{2784}            \\
ta09     & \textbf{1274}           &  & ta29       & 1625                     &  & ta49       & 1960                     &  & ta69       & \textbf{3071}            \\
ta10     & \textbf{1241}           &  & ta30       & 1584                     &  & ta50       & 1923                     &  & ta70       & \textbf{2995}            \\ \hline
ta11     & \textbf{1357}           &  & ta31       & \textbf{1764}            &  & ta51       & \textbf{2760}            &  & ta71       & \textbf{5464}            \\
ta12     & \textbf{1367}           &  & ta32       & 1784                     &  & ta52       & \textbf{2756}            &  & ta72       & \textbf{5181}            \\
ta13     & \textbf{1343}           &  & ta33       & 1791                     &  & ta53       & \textbf{2717}            &  & ta73       & \textbf{5568}            \\
ta14     & \textbf{1345}           &  & ta34       & \textbf{1828}            &  & ta54       & \textbf{2839}            &  & ta74       & \textbf{5339}            \\
ta15     & \textbf{1339}           &  & ta35       & \textbf{2007}            &  & ta55       & \textbf{2679}            &  & ta75       & \textbf{5392}            \\
ta16     & \textbf{1360}           &  & ta36       & \textbf{1819}            &  & ta56       & \textbf{2781}            &  & ta76       & \textbf{5342}            \\
ta17     & \textbf{1462}           &  & ta37       & \textbf{1771}            &  & ta57       & \textbf{2943}            &  & ta77       & \textbf{5436}            \\
ta18     & 1396                    &  & ta38       & \textbf{1673}            &  & ta58       & \textbf{2885}            &  & ta78       & \textbf{5394}            \\
ta19     & \textbf{1332}           &  & ta39       & \textbf{1795}            &  & ta59       & \textbf{2655}            &  & ta79       & \textbf{5358}            \\
ta20     & \textbf{1348}           &  & ta40       & 1670                     &  & ta60       & \textbf{2723}            &  & ta80       & \textbf{5183}            \\ \hline
\end{tabular}
\end{table}

\begin{table}[H]
\caption{Best-Known solutions (BKS) for \citet{vallada2015new} Flow Shop Scheduling instances with 10, 30, and 50 jobs, respectively. Values in \textbf{bold} indicate optimal solutions. }
\renewcommand{\arraystretch}{0.8}
\centering
\footnotesize
\begin{tabular}{cccccccc}
\hline
\multicolumn{2}{c}{VFR10}     &  & \multicolumn{2}{c}{VFR20} &  & \multicolumn{2}{c}{VFR50} \\ \cline{1-2} \cline{4-5} \cline{7-8} 
instance      & BKS           &  & instance         & BKS    &  & instance         & BKS    \\ \hline
VFR10\_10\_1  & \textbf{1051} &  & VFR30\_10\_1     & 1944   &  & VFR50\_10\_1     & 2926   \\
VFR10\_10\_2  & \textbf{1037} &  & VFR30\_10\_2     & 2098   &  & VFR50\_10\_2     & 3035   \\
VFR10\_10\_3  & \textbf{1071} &  & VFR30\_10\_3     & 2077   &  & VFR50\_10\_3     & 3019   \\
VFR10\_10\_4  & \textbf{984}  &  & VFR30\_10\_4     & 1945   &  & VFR50\_10\_4     & 3003   \\
VFR10\_10\_5  & \textbf{1060} &  & VFR30\_10\_5     & 2023   &  & VFR50\_10\_5     & 3253   \\
VFR10\_10\_6  & \textbf{1011} &  & VFR30\_10\_6     & 2043   &  & VFR50\_10\_6     & 3149   \\
VFR10\_10\_7  & \textbf{1026} &  & VFR30\_10\_7     & 1967   &  & VFR50\_10\_7     & 2842   \\
VFR10\_10\_8  & \textbf{1061} &  & VFR30\_10\_8     & 1896   &  & VFR50\_10\_8     & 3072   \\
VFR10\_10\_9  & \textbf{1020} &  & VFR30\_10\_9     & 1908   &  & VFR50\_10\_9     & 3022   \\
VFR10\_10\_10 & \textbf{1026} &  & VFR30\_10\_10    & 1915   &  & VFR50\_10\_10    & 3056   \\ \hline
VFR10\_20\_1  & \textbf{1568} &  & VFR30\_20\_1     & 2643   &  & VFR50\_20\_1     & 3693   \\
VFR10\_20\_2  & \textbf{1686} &  & VFR30\_20\_2     & 2835   &  & VFR50\_20\_2     & 3719   \\
VFR10\_20\_3  & \textbf{1640} &  & VFR30\_20\_3     & 2783   &  & VFR50\_20\_3     & 3784   \\
VFR10\_20\_4  & \textbf{1567} &  & VFR30\_20\_4     & 2680   &  & VFR50\_20\_4     & 3709   \\
VFR10\_20\_5  & \textbf{1570} &  & VFR30\_20\_5     & 2672   &  & VFR50\_20\_5     & 3632   \\
VFR10\_20\_6  & \textbf{1755} &  & VFR30\_20\_6     & 2715   &  & VFR50\_20\_6     & 3795   \\
VFR10\_20\_7  & \textbf{1633} &  & VFR30\_20\_7     & 2712   &  & VFR50\_20\_7     & 3696   \\
VFR10\_20\_8  & 1575          &  & VFR30\_20\_8     & 2812   &  & VFR50\_20\_8     & 3783   \\
VFR10\_20\_9  & \textbf{1666} &  & VFR30\_20\_9     & 2795   &  & VFR50\_20\_9     & 3816   \\
VFR10\_20\_10 & \textbf{1609} &  & VFR30\_20\_10    & 2805   &  & VFR50\_20\_10    & 3769   \\ \hline
\end{tabular}
\end{table}

\begin{table}[H]
\caption{Best-Known solutions (BKS) for \citet{hurink1994tabu} Flexible Job Shop Scheduling instances edata, vdata, rdata, respectively. Values in \textbf{bold} indicate optimal solutions.}
\centering
\footnotesize
\begin{tabular}{cccccccccccccc}
\hline
\multicolumn{4}{c}{Hurink   - edata}                &  & \multicolumn{4}{c}{Hurink - rdata}                  &  & \multicolumn{4}{c}{Hurink - vdata}                  \\ \cline{1-4} \cline{6-9} \cline{11-14} 
instance & BKS           & instance & BKS           &  & instance & BKS           & instance & BKS           &  & instance & BKS           & instance & BKS           \\ \hline
abz5     & \textbf{1167} & la21     & \textbf{1009} &  & abz5     & \textbf{954}  & la21     & 835           &  & abz5     & \textbf{859}  & la21     & 803           \\
abz6     & \textbf{925}  & la22     & \textbf{880}  &  & abz6     & \textbf{807}  & la22     & 760           &  & abz6     & \textbf{742}  & la22     & 736           \\
abz7     & 622           & la23     & \textbf{950}  &  & abz7     & 544           & la23     & 842           &  & abz7     & 495           & la23     & 811           \\
abz8     & 644           & la24     & \textbf{908}  &  & abz8     & 555           & la24     & 804           &  & abz8     & 509           & la24     & 775           \\
abz9     & 648           & la25     & \textbf{936}  &  & abz9     & 545           & la25     & 787           &  & abz9     & 500           & la25     & 756           \\
car1     & \textbf{6176} & la26     & \textbf{1106} &  & car1     & \textbf{5034} & la26     & 1061          &  & car1     & 5006          & la26     & 1054          \\
car2     & \textbf{6327} & la27     & 1181          &  & car2     & \textbf{5985} & la27     & 1056          &  & car2     & 5929          & la27     & \textbf{1084} \\
car3     & \textbf{6856} & la28     & 1147          &  & car3     & 5623          & la28     & 1080          &  & car3     & 5599          & la28     & 1070          \\
car4     & \textbf{7789} & la29     & 1107          &  & car4     & 6514          & la29     & 998           &  & car4     & 6514          & la29     & 994           \\
car5     & \textbf{7229} & la30     & 1204          &  & car5     & \textbf{5615} & la30     & 1078          &  & car5     & 4919          & la30     & 1069          \\
car6     & \textbf{7990} & la31     & \textbf{1532} &  & car6     & \textbf{6147} & la31     & 1521          &  & car6     & \textbf{5486} & la31     & \textbf{1520} \\
car7     & \textbf{6123} & la32     & \textbf{1698} &  & car7     & \textbf{4425} & la32     & 1659          &  & car7     & \textbf{4281} & la32     & 1658          \\
car8     & \textbf{7689} & la33     & \textbf{1547} &  & car8     & \textbf{5692} & la33     & 1499          &  & car8     & \textbf{4613} & la33     & \textbf{1497} \\
la01     & \textbf{609}  & la34     & \textbf{1599} &  & la01     & \textbf{570}  & la34     & 1536          &  & la01     & \textbf{570}  & la34     & \textbf{1535} \\
la02     & \textbf{655}  & la35     & \textbf{1736} &  & la02     & \textbf{529}  & la35     & 1550          &  & la02     & \textbf{529}  & la35     & \textbf{1549} \\
la03     & \textbf{550}  & la36     & \textbf{1160} &  & la03     & \textbf{477}  & la36     & \textbf{1023} &  & la03     & \textbf{477}  & la36     & \textbf{948}  \\
la04     & \textbf{568}  & la37     & \textbf{1397} &  & la04     & \textbf{502}  & la37     & 1066          &  & la04     & \textbf{502}  & la37     & \textbf{986}  \\
la05     & \textbf{503}  & la38     & \textbf{1141} &  & la05     & \textbf{457}  & la38     & \textbf{954}  &  & la05     & \textbf{457}  & la38     & \textbf{943}  \\
la06     & \textbf{833}  & la39     & \textbf{1184} &  & la06     & \textbf{799}  & la39     & \textbf{1011} &  & la06     & \textbf{799}  & la39     & \textbf{922}  \\
la07     & \textbf{762}  & la40     & \textbf{1144} &  & la07     & 749           & la40     & \textbf{955}  &  & la07     & \textbf{749}  & la40     & \textbf{955}  \\
la08     & \textbf{845}  & mt06     & \textbf{55}   &  & la08     & \textbf{765}  & mt06     & \textbf{47}   &  & la08     & \textbf{765}  & mt06     & \textbf{47}   \\
la09     & \textbf{878}  & mt10     & \textbf{871}  &  & la09     & \textbf{853}  & mt10     & \textbf{686}  &  & la09     & \textbf{853}  & mt10     & \textbf{655}  \\
la10     & \textbf{866}  & mt20     & \textbf{1088} &  & la10     & \textbf{804}  & mt20     & 1022          &  & la10     & \textbf{804}  & mt20     & \textbf{1022} \\
la11     & \textbf{1103} & orb1     & \textbf{977}  &  & la11     & \textbf{1071} & orb1     & \textbf{746}  &  & la11     & \textbf{1071} & orb1     & \textbf{695}  \\
la12     & \textbf{960}  & orb2     & \textbf{865}  &  & la12     & \textbf{936}  & orb2     & \textbf{696}  &  & la12     & \textbf{936}  & orb2     & \textbf{620}  \\
la13     & \textbf{1053} & orb3     & \textbf{951}  &  & la13     & \textbf{1038} & orb3     & \textbf{712}  &  & la13     & \textbf{1038} & orb3     & \textbf{648}  \\
la14     & \textbf{1123} & orb4     & \textbf{984}  &  & la14     & \textbf{1070} & orb4     & \textbf{753}  &  & la14     & \textbf{1070} & orb4     & \textbf{753}  \\
la15     & \textbf{1111} & orb5     & \textbf{842}  &  & la15     & 1089          & orb5     & \textbf{639}  &  & la15     & \textbf{1089} & orb5     & \textbf{584}  \\
la16     & \textbf{892}  & orb6     & \textbf{958}  &  & la16     & \textbf{717}  & orb6     & \textbf{754}  &  & la16     & \textbf{717}  & orb6     & \textbf{715}  \\
la17     & \textbf{707}  & orb7     & \textbf{389}  &  & la17     & \textbf{646}  & orb7     & \textbf{302}  &  & la17     & \textbf{646}  & orb7     & \textbf{275}  \\
la18     & \textbf{842}  & orb8     & \textbf{894}  &  & la18     & \textbf{666}  & orb8     & \textbf{639}  &  & la18     & \textbf{663}  & orb8     & \textbf{573}  \\
la19     & \textbf{796}  & orb9     & \textbf{933}  &  & la19     & \textbf{700}  & orb9     & \textbf{694}  &  & la19     & \textbf{617}  & orb9     & \textbf{659}  \\
la20     & \textbf{857}  & orb10    & \textbf{933}  &  & la20     & \textbf{756}  & orb10    & \textbf{742}  &  & la20     & \textbf{756}  & orb10    & \textbf{681}  \\ \hline
\end{tabular}
\end{table}

\begin{table}[H]
\centering
\footnotesize
\caption{Best-Known solutions (BKS) for \citet{brandimarte1993routing} Flexible Job Shop Scheduling instances 1-10, respectively. Values in \textbf{bold} indicate optimal solutions.}
\begin{tabular}{cc}
\hline
\multicolumn{2}{c}{Brandimarte 1-10} \\ \hline
instance        & BKS                \\ \hline
Mk01           & \textbf{40}        \\
Mk02           & 26                 \\
Mk03           & \textbf{204}       \\
Mk04           & \textbf{60}        \\
Mk05           & 172                \\
Mk06           & 58                 \\
Mk07           & 139                \\
Mk08           & \textbf{523}       \\
Mk09           & \textbf{307}       \\
Mk10           & 197                \\ \hline
\end{tabular}
\end{table}

\newpage
\section{Extended Performance Results}
\label{appendix:extended_results}

\begin{table}[H]
\caption{Performance Results JSP (\citet{taillard1993benchmarks} instances: 15x15 to 30x15)}
\centering
\scriptsize
\begin{tabular}{lcclcclcclcclcclcclcclcc}

\hline
                    & \multicolumn{2}{c}{TA   15x15} &  & \multicolumn{2}{c}{TA 20x15} &  & \multicolumn{2}{c}{TA 20x20} &  & \multicolumn{2}{c}{TA 30x15} \\ 
                    \cline{2-3} \cline{5-6} \cline{8-9} \cline{11-12}
                    & obj            & gap           &  & obj           & gap          &  & obj           & gap          &  & obj           & gap         \\ \hline
CP-SAT              & 1229           & 0,01          &  & 1373,9        & 0,66         &  & 1628,8        & 0,71         &  & 1813,1        & 1,28        \\
MILP                & 1232,2         & 0,27          &  & 1422,2        & 4,20         &   & 1664,2        & 2,90         &  & 1958,1        & 9,38       \\
SPT                 & 1546,1         & 25,81         &  & 1813,5        & 32,87        &  & 2067          & 27,81        &  & 2419,3        & 35,14       \\
MWR                 & 1464,3         & 19,16         &  & 1683,6        & 23,35        &  & 1969,8        & 21,80        &  & 2214,8        & 23,72       \\
MOR                 & 1481,3         & 20,54         &  & 1686,7        & 23,58        &  & 1968,3        & 21,70        &  & 2195,8        & 22,66       \\
GA 100x100          & 1583,8         & 28,88         &  & 1847,1        & 35,33        &  & 2304,2        & 42,47        &  & 2493          & 39,26        \\
L2D 5x5 greedy      & 1589,1         & 29,31         &  & 1869,9        & 37,00        &  & 2271,8        & 40,47        &  & 2436,2        & 36,09        \\
L2D 5x5 sample      & 1760,2         & 43,23         &  & 2083,7        & 52,66        &  & 2610,6        & 61,42        &  & 2852,9        & 59,36       \\
L2D 10x10 greedy    & 1584,8         & 28,96         &  & 1759          & 28,87        &  & 2090,8        & 29,28        &  & 2315,2        & 29,33        \\
L2D 10x10 sample    & 1623,6         & 32,12         &  & 1900,3        & 39,23        &  & 2340,2        & 44,70        &  & 2564,7        & 43,26        \\
L2D 20x20 greedy    & 1522,3         & 23,88         &  & 1813,4        & 32,86        &  & 2130,2        & 31,71        &  & 2348,7        & 31,20        \\
L2D 20x20 sample    & 1543,8         & 25,62         &  & 1779,7        & 30,39        &  & 2134,2        & 31,96        &  & 2396          & 33,84        \\
FJSP-DRL 5x5   greedy   & 1529,8         & 24,49         &  & 1802,5        & 32,06        &  & 2068,8        & 27,92        &  & 2383,8        & 33,16        \\
FJSP-DRL 5x5 sample     & 1425,9         & 16,03         &  & 1660,7        & 21,67        &  & 1934,7        & 19,63        &  & 2203,6        & 23,09       \\
FJSP-DRL 10x10 greedy   & 1560,7         & 27,00         &  & 1743,6        & 27,75        &  & 2035,3        & 25,85        &  & 2253,2        & 25,86        \\
FJSP-DRL 10x10 sample   & 1395,1         & 13,52         &  & 1613,8        & 18,24        &  & 1897,9        & 17,35        &  & 2132,8        & 19,14        \\
FJSP-DRL 20x20 greedy   & 1491           & 21,33         &  & 1715,9        & 25,72        &  & 1994,8        & 23,34        &  & 2315          & 29,32       \\
FJSP-DRL 20x20 sample   & 1420,9         & 15,62         &  & 1645,2        & 20,54        &  & 1933          & 19,52        &  & 2183,1        & 21,95        \\
DANIEL 5x5 greedy   & 1453,1         & 18,24         &  & 1669,7        & 22,33        &  & 1937,3        & 19,79        &  & 2188,7        & 22,26        \\
DANIEL 5x5 sample   & 1407,1         & 14,50         &  & 1593,9        & 16,78        &  & 1858,5        & 14,91        &  & 2121,5        & 18,51       \\
DANIEL 10x10 greedy & 1462,6         & 19,02         &  & 1666,7        & 22,11        &  & 1908,9        & 18,03        &  & 2178,2        & 21,67        \\
DANIEL 10x10 sample & 1391,4         & 13,22         &  & 1602,2        & 17,39        &  & 1831,6        & 13,25        &  & 2098,1        & 17,20        \\
DANIEL 20x20 greedy & 1454,6         & 18,37         &  & 1676,2        & 22,81        &  & 1936,8        & 19,76        &  & 2196,3        & 22,68        \\
DANIEL 20x20 sample & 1417           & 15,31         &  & 1607,5        & 17,77        &  & 1860          & 15,01        &  & 2102,3        & 17,43       \\ \hline
BKS  & \multicolumn{2}{c}{1228,9}     &  & \multicolumn{2}{c}{1364,9}   &  & \multicolumn{2}{c}{1617,3}   &  & \multicolumn{2}{c}{1790,2}   \\ \hline
\end{tabular}
\end{table}

\begin{table}[H]
\caption{Performance Results JSP (\citet{taillard1993benchmarks} instances: 30x20 to 100x20)}
\centering
\scriptsize
\begin{tabular}{lcclcclcclcclcclcclcclcc}
\hline
                    & \multicolumn{2}{c}{TA 30x20} &  & \multicolumn{2}{c}{TA 50x15} &  & \multicolumn{2}{c}{TA 50x20} &  & \multicolumn{2}{c}{TA 100x20} \\ 
                    \cline{2-3} \cline{5-6} \cline{8-9} \cline{11-12}

                    & obj            & gap           &  & obj           & gap          &  & obj           & gap          &  & obj           & gap         \\ \hline

CP-SAT             & 2009,9        & 3,15         &  & 2773,8        & 0,00         &  & 2850,3        & 0,23         &  & 5365,7        & 0,00          \\
MILP               & 2172,9        & 11,52        &  &      -         &      -       &  &           -    &      -        &  & 5365,7        & 0,00          \\
SPT                & 2619,1        & 34,42        &  & 3441          & 24,05        &  & 3570,8        & 25,56        &  & 6139          & 14,41         \\
MWR                & 2439          & 25,17        &  & 3240          & 16,81        &  & 3352,8        & 17,89        &  & 5812,2        & 8,32          \\
MOR                & 2433,6        & 24,90        &  & 3254,5        & 17,33        &  & 3346,9        & 17,69        &  & 5856,9        & 9,15          \\
GA 100x100         & 2925,6        & 50,15        &  & 3439,5        & 24,00        &  & 3965,9        & 39,45        &  & 6251          & 16,50         \\
L2D 5x5 greedy     & 2927,6        & 50,25        &  & 3631,2        & 30,91        &  & 4424,5        & 55,58        &  & 7035,2        & 31,11         \\
L2D 5x5 sample     & 3430          & 76,03        &  & 4164,2        & 50,13        &  & 4815,7        & 69,33        &  & 8255,2        & 53,85         \\
L2D 10x10 greedy   & 2584,3        & 32,63        &  & 3366,5        & 21,37        &  & 3538,1        & 24,41        &  & 6105,7        & 13,79         \\
L2D 10x10 sample   & 2997,3        & 53,83        &  & 3780,9        & 36,31        &  & 4262,5        & 49,88        &  & 7328,3        & 36,58         \\
L2D 20x20 greedy   & 2585,6        & 32,70        &  & 3364,9        & 21,31        &  & 3546,9        & 24,72        &  & 6129,9        & 14,24         \\
L2D 20x20 sample   & 2749,3        & 41,10        &  & 3488,8        & 25,78        &  & 3821          & 34,36        &  & 6668,9        & 24,29         \\
FJSP-DRL 5x5   greedy   & 2613,1        & 34,11        &  & 3438,9        & 23,98        &  & 3587,1        & 26,13        &  & 6226,2        & 16,04         \\
FJSP-DRL 5x5 sample     & 2470,1        & 26,77        &  & 3205,9        & 15,58        &  & 3383,3        & 18,97        &  & 5940,6        & 10,71         \\
FJSP-DRL 10x10 greedy   & 2531,8        & 29,94        &  & 3313,9        & 19,47        &  & 3443,1        & 21,07        &  & 5862,9        & 9,27          \\
FJSP-DRL 10x10 sample   & 2385,5        & 22,43        &  & 3118,3        & 12,42        &  & 3289,4        & 15,67        &  & 5723,6        & 6,67          \\
FJSP-DRL 20x20 greedy   & 2540          & 30,36        &  & 3355,3        & 20,96        &  & 3526,5        & 24,00        &  & 6247,1        & 16,43         \\
FJSP-DRL 20x20 sample   & 2442,7        & 25,36        &  & 3207,5        & 15,64        &  & 3372,2        & 18,58        &  & 5932,7        & 10,57         \\
DANIEL 5x5 greedy   & 2421,9        & 24,30        &  & 3204,2        & 15,52        &  & 3308          & 16,32        &  & 5784,9        & 7,81          \\
DANIEL 5x5 sample   & 2337,7        & 19,97        &  & 3142,9        & 13,31        &  & 3232,8        & 13,67        &  & 5705,9        & 6,34          \\
DANIEL 10x10 greedy & 2401,2        & 23,23        &  & 3183,2        & 14,76        &  & 3298          & 15,97        &  & 5758,8        & 7,33          \\
DANIEL 10x10 sample & 2317,7        & 18,95        &  & 3126,8        & 12,73        &  & 3217          & 13,12        &  & 5681,2        & 5,88          \\
DANIEL 20x20 greedy & 2407,5        & 23,56        &  & 3205,9        & 15,58        &  & 3304,4        & 16,19        &  & 5740,5        & 6,99          \\
DANIEL 20x20 sample & 2315,5        & 18,84        &  & 3122,7        & 12,58        &  & 3220,7        & 13,25        &  & 5644,4        & 5,19          \\ \hline
BKS  & \multicolumn{2}{c}{1948,5}   &  & \multicolumn{2}{c}{2773,8}   &  & \multicolumn{2}{c}{2843,9}   &  & \multicolumn{2}{c}{5365,7}    \\ \hline
\end{tabular}
\end{table}

\begin{table}[H]
\caption{Performance Results FSP (\citet{vallada2015new} instances: 10x10 to 50x20)}
\centering
\renewcommand{\arraystretch}{1.1}
\scriptsize
\begin{tabular}{lc@{\hskip 0.07in}c@{\hskip 0.01in}lc@{\hskip 0.07in}c@{\hskip 0.01in}lc@{\hskip 0.07in}c@{\hskip 0.01in}lc@{\hskip 0.07in}c@{\hskip 0.01in}lc@{\hskip 0.07in}c@{\hskip 0.01in}lc@{\hskip 0.07in}c@{\hskip 0.01in}lc@{\hskip 0.07in}c@{\hskip 0.01in}lcc}

\hline
                    & \multicolumn{2}{c}{VRF 10x10} &  & \multicolumn{2}{c}{VRF 10x20} &  & \multicolumn{2}{c}{VRF 30x10} &  & \multicolumn{2}{c}{VRF 30x20} &  & \multicolumn{2}{c}{VRF 50x10} &  & \multicolumn{2}{c}{VRF 50x20} \\ \cline{2-3} \cline{5-6} \cline{8-9} \cline{11-12} \cline{14-15} \cline{17-18} 
                    & obj            & gap          &  & obj            & gap          &  & obj           & gap           &  & obj           & gap           &  & obj           & gap           &  & obj            & gap          \\ \hline
CP-SAT              & 1034,7         & 0,00         &  & 1626,9         & 0,00         &  & 2015,6        & 1,72          &  & 2841,2        & 3,50          &  & 3133,4        & 3,15          &  & 4037,1         & 7,96         \\
MILP                & 1034,7         & 0,00         &  & 1634,3         & 0,45         &  & 2254,8        & 13,79         &  & 3298,3        & 20,15         &  & 4671,2        & 53,77         &  & -               & -             \\
SPT                 & 1216,3         & 17,55        &  & 1880,9         & 15,61        &  & 2320,6        & 17,11         &  & 3197,3        & 16,47         &  & 3450,1        & 13,58         &  & 4434,3         & 18,58        \\
MRW                 & 1243,9         & 20,22        &  & 1936,6         & 19,04        &  & 2570,7        & 29,73         &  & 3462,7        & 26,14         &  & 3830,8        & 26,11         &  & 4767,1         & 27,48        \\
MOR                 & 1263,9         & 22,15        &  & 1971,9         & 21,21        &  & 2515,6        & 26,95         &  & 3421,9        & 24,65         &  & 3798          & 25,03         &  & 4758,1         & 27,24        \\
GA 100x100          & 1108           & 7,08         &  & 1750,9         & 7,62         &  & 2307,6        & 16,45         &  & 3184,2        & 15,99         &  & 3630,8        & 19,52         &  & 4459,3         & 19,25        \\
L2D 5x5 greedy      & 1523,6         & 47,25        &  & 2287,1         & 40,58        &  & 3984,6        & 101,08        &  & 4364,8        & 59,00         &  & 4845,7        & 59,52         &  & 6045,2         & 61,65        \\
L2D 5x5 sample      & 1462,4         & 41,34        &  & 2539,9         & 56,12        &  & 4336,7        & 118,85        &  & 6625,2        & 141,34        &  & 7738,3        & 154,74        &  & 10834,7        & 189,73       \\
L2D 10x10 greedy    & 1360           & 31,44        &  & 2072,3         & 27,38        &  & 2596          & 31,01         &  & 3704          & 34,93         &  & 4366,3        & 43,74         &  & 5315,4         & 42,14        \\
L2D 10x10 sample    & 1331,5         & 28,68        &  & 2351,5         & 44,54        &  & 3759,7        & 89,73         &  & 6015          & 119,11        &  & 6817          & 124,41        &  & 10523,7        & 181,41       \\
L2D 20x20 greedy    & 1306,6         & 26,28        &  & 1921,4         & 18,10        &  & 2602,5        & 31,33         &  & 3460,1        & 26,04         &  & 3853          & 26,84         &  & 4847,7         & 29,63        \\
L2D 20x20 sample    & 1237           & 19,55        &  & 1966,7         & 20,89        &  & 2954,3        & 49,09         &  & 4059,9        & 47,89         &  & 4899,9        & 61,30         &  & 6363           & 70,15        \\
FJSP-DRL 5x5   greedy   & 1243,9         & 20,22        &  & 1996,4         & 22,71        &  & 2641,8        & 33,32         &  & 3483,9        & 26,91         &  & 4019,7        & 32,33         &  & 4907,8         & 31,24        \\
FJSP-DRL 5x5 sample     & 1149,6         & 11,10        &  & 1784,1         & 9,66         &  & 2334,2        & 17,79         &  & 3208,9        & 16,89         &  & 3531,9        & 16,27         &  & 4495,2         & 20,21        \\
FJSP-DRL 10x10 greedy   & 1253,9         & 21,18        &  & 1976,7         & 21,50        &  & 2669          & 34,69         &  & 3520,8        & 28,25         &  & 4047,9        & 33,26         &  & 4808,8         & 28,59        \\
FJSP-DRL 10x10 sample   & 1156           & 11,72        &  & 1817,9         & 11,74        &  & 2385,1        & 20,36         &  & 3232,7        & 17,76         &  & 3613,2        & 18,95         &  & 4521,9         & 20,92        \\
FJSP-DRL 20x20 greedy   & 1236,4         & 19,49        &  & 1949,4         & 19,82        &  & 2545,3        & 28,45         &  & 3426,3        & 24,81         &  & 3825,4        & 25,93         &  & 4700,8         & 25,70        \\
FJSP-DRL 20x20 sample   & 1151           & 11,24        &  & 1795,5         & 10,36        &  & 2331,8        & 17,67         &  & 3184,8        & 16,01         &  & 3551,5        & 16,91         &  & 4478,3         & 19,75        \\
DANIEL 5x5 greedy   & 1247,3         & 20,55        &  & 1960,4         & 20,50        &  & 2532,5        & 27,80         &  & 3425,5        & 24,78         &  & 3731,4        & 22,84         &  & 4703,6         & 25,78        \\
DANIEL 5x5 sample   & 1205,8         & 16,54        &  & 1898,7         & 16,71        &  & 2415,3        & 21,89         &  & 3299,9        & 20,21         &  & 3624,3        & 19,31         &  & 4560,4         & 21,95        \\
DANIEL 10x10 greedy & 1227,8         & 18,66        &  & 1943,9         & 19,48        &  & 2427,9        & 22,52         &  & 3346,8        & 21,91         &  & 3618          & 19,10         &  & 4560           & 21,94        \\
DANIEL 10x10 sample & 1193           & 15,30        &  & 1888,7         & 16,09        &  & 2347,1        & 18,44         &  & 3237,1        & 17,92         &  & 3540,7        & 16,56         &  & 4447,5         & 18,93        \\
DANIEL 20x20 greedy & 1256,6         & 21,45        &  & 1940,8         & 19,29        &  & 2431,3        & 22,69         &  & 3367,8        & 22,68         &  & 3581,2        & 17,89         &  & 4501,6         & 20,38        \\
DANIEL 20x20 sample & 1228,5         & 18,73        &  & 1863,4         & 14,54        &  & 2346,4        & 18,41         &  & 3224,2        & 17,45         &  & 3473,1        & 14,33         &  & 4368,8         & 16,83        \\ \hline
BKS    & \multicolumn{2}{c}{1034,7}    &  & \multicolumn{2}{c}{1626,9}    &  & \multicolumn{2}{c}{1981,6}    &  & \multicolumn{2}{c}{2745,2}    &  & \multicolumn{2}{c}{3037,7}    &  & \multicolumn{2}{c}{3739,6}    \\ \hline
\end{tabular}
\end{table}

\begin{table}[H]
\centering
\caption{Results FJSP (\citet{brandimarte1993routing} \& \citet{hurink1994tabu} instances)}
\small
\begin{tabular}{lcclcclcclcc}
\hline
                    & \multicolumn{2}{c}{Brandimarte} &  & \multicolumn{2}{c}{Hurink (edata)} & \multicolumn{1}{c}{} & \multicolumn{2}{c}{Hurink (rdata)} &  & \multicolumn{2}{c}{Hurink (vdata)} \\ \cline{2-3} \cline{5-6} \cline{8-9} \cline{11-12} 
                    & obj            & gap            &  & obj               & gap             &                      & obj               & gap             &  & obj               & gap             \\ \hline
CP-SAT              & 173,7          & 0,64           &  & 1697,1            & 0,00            &                      & 1427,7            & 0,09            &  & 1367,2            & 0,09            \\
MILP                & 180,6          & 4,63           &  & 1711,6            & 0,86            &                      & 1448,7            & 1,56            &  & 1394,4            & 2,08            \\
SPT                 & 268            & 55,27          &  & 2053,0            & 20,97           &                      & 1813,8            & 27,16           &  & 1622,0            & 18,74           \\
MWR - SPT           & 245            & 41,95          &  & 2053,9            & 21,03           &                      & 1690,9            & 18,55           &  & 1439,0            & 5,35            \\
MWR - EET           & 186,1          & 7,82           &  & 2052,1            & 20,92           &                      & 1692,9            & 18,69           &  & 1447,0            & 5,93            \\
MOR - SPT           & 244,9          & 41,89          &  & 2096,8            & 23,56           &                      & 1719,6            & 20,56           &  & 1519,7            & 11,26           \\
MOR - EET            & 186,1          & 7,82           &  & 2052,0            & 20,92           &                      & 1692,9            & 18,68           &  & 1447,1            & 5,94            \\
GA 100x100          & 207,4          & 20,16          &  & 1942,2            & 14,44           &                      & 1840,5            & 29,03           &  & 1750,6            & 28,16           \\
FJSP-DRL 5x5   greedy   & 221            & 28,04          &  & 2125,3            & 25,24           &                      & 1679,3            & 17,73           &  & 1544,2            & 13,05           \\
FJSP-DRL 5x5 sample     & 197,9          & 14,66          &  & 1893,6            & 11,58           &                      & 1574,4            & 10,38           &  & 1417,8            & 3,80            \\
FJSP-DRL 10x10 greedy   & 206,1          & 19,41          &  & 2103,9            & 23,97           &                      & 1665,2            & 16,74           &  & 1462,5            & 7,07            \\
FJSP-DRL 10x10 sample   & 192,4          & 11,47          &  & 1904,6            & 12,23           &                      & 1526,9            & 7,05            &  & 1387,8            & 1,60            \\
FJSP-DRL 20x20 greedy   & 237,3          & 37,49          &  & 2130,8            & 25,56           &                      & 1733,0            & 21,50           &  & 1524,2            & 11,59           \\
FJSP-DRL 20x20 sample   & 200,7          & 16,28          &  & 1899,1            & 11,91           &                      & 1561,8            & 9,50            &  & 1410,5            & 3,27            \\
DANIEL 5x5 greedy   & 185,3          & 7,36           &  & 2032,3            & 19,76           &                      & 1617,0            & 13,36           &  & 1436,3            & 5,15            \\
DANIEL 5x5 sample   & 181,4          & 5,10           &  & 1938,6            & 14,23           &                      & 1525,0            & 6,91            &  & 1389,3            & 1,71            \\
DANIEL 10x10 greedy & 185            & 7,18           &  & 2010,1            & 18,45           &                      & 1647,9            & 15,53           &  & 1442,5            & 5,61            \\
DANIEL 10x10 sample & 181,2          & 4,98           &  & 1918,8            & 13,07           &                      & 1542,0            & 8,11            &  & 1385,8            & 1,46            \\
DANIEL 20x20 greedy & 186,9          & 8,29           &  & 2060,9            & 21,44           &                      & 1674,6            & 17,40           &  & 1461,4            & 6,99            \\
DANIEL 20x20 sample & 182,4          & 5,68           &  & 1970,3            & 16,11           &                      & 1562,1            & 9,51            &  & 1396,5            & 2,23            \\ \hline
BKS               & \multicolumn{2}{c}{172,6}                          &  & \multicolumn{2}{c}{1697,0}                 &                      & \multicolumn{2}{c}{1426,4}                 &  & \multicolumn{2}{c}{1365,9} \\ \hline
\end{tabular}
\end{table}

\end{APPENDICES}


\bibliographystyle{informs2014} 
\bibliography{references}

\end{document}